\documentclass{article}
\usepackage{arxiv}

\usepackage[utf8]{inputenc} 
\usepackage[T1]{fontenc}    
\usepackage{hyperref}       
\usepackage{url}            
\usepackage{booktabs}       
\usepackage{amsfonts}       
\usepackage{nicefrac}       
\usepackage{microtype}      
\usepackage{lipsum}		
\usepackage{graphicx}
\usepackage{doi}
\usepackage{placeins}
\usepackage{times}
\usepackage{epsfig}
\usepackage[numbers]{natbib}

\title{Deep radiomic signature with immune cell markers predicts the survival of glioma patients}

\author{ Ahmad Chaddad $^{1,4}$*, Paul Daniel $^{2}$, Mingli Zhang $^{5}$, Saima Rathore $^{6}$,Paul Sargos $^{4}$, Christian Desrosiers$^{3}$,\\ 
\textbf{Tamim Niazi} $^{4}$ \\
$^{1}$ \quad School of Artificial intelligence, Guilin Universiy of Electronic Technology, China \\
$^{2}$ \quad Hudson Institute of Medical Research, Australia \\
$^{3}$ \quad The Laboratory for Imagery, Vision and Artificial Intelligence, Canada \\
$^{4}$ \quad Lady Davis Institute for Medical Research, McGill University, Canada \\
$^{5}$ \quad Montreal Neurological Institute, McGill University, Canada\\
$^{6}$ \quad Perelman School of Medicine, University of Pennsylvania,USA \\
email:ahmad8chaddad@gmail.com, ahmadchaddad@guet.edu.cn}

\begin{document}
\maketitle

\begin{abstract}
Imaging biomarkers offer a non-invasive way to predict the response of immunotherapy prior to treatment. In this work, we propose a novel type of deep radiomic features (DRFs) computed from a convolutional neural network (CNN), which capture tumor characteristics related to immune cell markers and overall survival. Our study uses four MRI sequences (T1-weighted, T1-weighted post-contrast, T2-weighted and FLAIR) with corresponding immune cell markers of 151 patients with brain tumor. The proposed method extracts a total of 180 DRFs by aggregating the activation maps of a pre-trained 3D-CNN within labeled tumor regions of MRI scans. These features offer a compact, yet powerful representation of regional texture encoding tissue heterogeneity. A comprehensive set of experiments is performed to assess the relationship between the proposed DRFs and immune cell markers, and measure their association with overall survival. Results show a high correlation between DRFs and various markers, as well as significant differences between patients grouped based on these markers. Moreover, combining DRFs, clinical features and immune cell markers as input to a random forest classifier helps discriminate between short and long survival outcomes, with AUC of 72\% and p=2.36$\times$10$^{-5}$. These results demonstrate the usefulness of proposed DRFs as non-invasive biomarker for predicting treatment response in patients with brain tumors.
\end{abstract}



\section{Introduction}

Gliomas are the most common tumors initiating in the brain. As defined by the World Health Organization (WHO) \cite{ref1}, gliomas can be classified into four grades (I, II, III or IV) by histopathological process depending on their aggressiveness. Grade I glioma represents non-invasive tumors, grade II/III corresponds to low/intermediate-grade gliomas also named lower grade glioma (LGG), and grade IV to aggressive malignant tumors called glioblastoma multiforme (GBM). GBM is the most deadly brain tumor with a median survival of 15 months \cite{ref2}. Most patients relapse within months, after which there are limited options for further treatment. Immunotherapy is a promising strategy for cancer treatment in which the patient's own immune system is used to eliminate cancer cells \cite{ref3}. Despite encouraging developments, predicting response to immunotherapy prior to treatment remains a challenging task for clinicians.

Intratumoral immune response was shown to be related to tumor progression and prognosis in gliomas \cite{ref4,ref5,ref6}. Various markers, such as the CD3 (cluster of differentiation 3) marker, were investigated for evaluating intratumoral immune response. CD3 is a protein complex composed of CD3G, CD3D, and CD3E chains that is considered a general marker of T-cells. Antitumor immune responses have been correlated with clinical response to the immune therapeutic \cite{ref7}, with tumor-infiltrating CD3 T-cells and dendritic cell therapy for GBM patients \cite{ref8}. Despite the high potential of immunothearpy, the prediction of response to immune therapeutics usually requires an invasive technique by either biopsy or surgery, which is costly and has inherent risks of complications. Hence, the development of reliable imaging biomarkers that can capture charcteristics of immune cells would provide a non-invasive alternative for improving the immune therapeutics process.

Recent work has examined the relationships between tumor imaging features (e.g., shape, texture, histogram), multi-omics (e.g., genomics, proteomics, transcriptomics) and clinical outcome \cite{ref9,ref10,ref11}. Radiomics is a computational approach that seeks to convert medical images into quantitative data \cite{ref12}. It has shown the ability to capture characteristics of tissue heterogeneity that are related to cellular and molecular properties \cite{ref13,ref14,ref15}. Radiomic methods are typically used to derive image biomarkers for computer-aided diagnosis (CAD), and to monitor patients in pre- and post-treatment. Investigating the link between radiomic features and intratumoral immune response may thus provide a non-invasive technique for evaluating the benefit of immune therapeutics in individual patients \cite{ref16}. In a recent study  \cite{ref17}, a structured approach is provided to decipher tumor characteristics and its immune environment. Further, the usefulness of radiomic signatures has been demonstrated in estimating CD8 cell count and predicting clinical outcomes of patients treated with immunotherapy \cite{ref18}. Another study \cite{ref19} presents available immunotherapy regimens for evaluating the anti-tumor and immune responses to immunotherapy in neuro-oncology applications. The ratio of tumor volume in T2-FLAIR scans relative to the volume of contrast enhancement was shown to be associated to outcome for the mesenchymal subtype of GBM \cite{ref20} which has a stronger immunological response compared to other GBM subtypes \cite{ref21}. 

So far, no investigation has explored the potential of radiomic analysis to predict immune cell response and its impact on survival outcome for patients with LGG and GBM. To bridge this gap, we propose a novel radiomic signature for LGG and GBM brain tumors, based on the activation maps of a pre-trained 3D convolutional neural network (CNN). Our model extends the work in \cite{ref22,ref32,ref38}, where the entropy of CNN activations was considered as measure of texture heterogeneity, and proposes a broader set of CNN-based features which capture important characteristics of tumor heterogeneity that are related to immune cell markers. Specifically, the major contributions of this work are as follows:
\begin{itemize} 
\item To our knowledge, this work is the first to encode CNN activation maps with standard radiomic functions for survival and immune cell marker prediction. The proposed Deep Radiomic Features (DRFs) offer a compact representation of image texture, which captures the heterogeneity of glioma tissues at different image scales.
\item We present a detailed evaluation of the proposed DRFs on one of the largest publicly-available brain tumor datasets. Our results show their relationship to several immune cell markers and to overall patient survival. When used as input to a random forest classifier, our DRFs lead to a significantly higher accuracy than clinical features and immune cell markers for predicting shorter versus longer survival groups.
\end{itemize}  

The rest of this paper is structured as follows. Section II describes the image data, clinical and immune cell markers, proposed deep radiomic features for predicting the immune cell scores and clinical outcomes of glioma patients. Section III provides experimental setup and results. Section IV discusses our findings. Finally, Section 5 concludes with a summary of our work’s main contributions and results.

\begin{figure}[t!]
    \centering
    \includegraphics[width=1\textwidth]{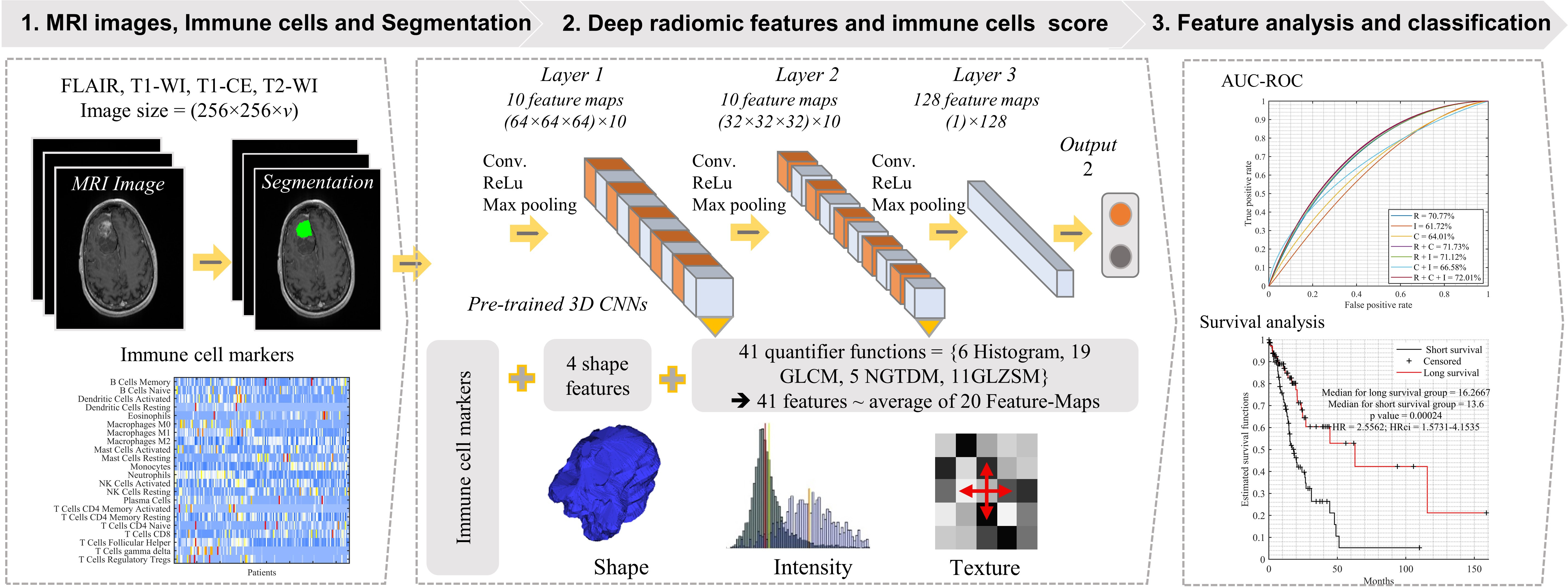}
    \caption{Deep radiomic pipeline for predicting immune cell markers and survival. 1) Image acquisition with manually segmented tumor masks. 2) Deep image feature extraction. 3) Predictive models and informatics analysis.}
    \label{fig1}
\end{figure}

\section{Materials and methods}

Figure \ref{fig1} shows the pipeline of the proposed deep radiomic model. Four MRI sequences (T1-WI, T1-CE, FLAIR and T2WI) are first acquired. Segmentation is then performed by labelling the tumor regions of interest (ROIs) in each scan. Thereafter, DRFs are extracted from 3D CNN activations corresponding to segmented ROIs. Finally, statistical and classification analyses are conducted to assess the relationship between DRFs and immune  cell markers and the DRFs' ability to predict the survival of glioma patients. The following subsections present each of these steps in greater detail. 

\subsection{Patient selection and image preprocessing}

For this study, we used pre-surgical images and immune cell markers data from 151 patients in the TCGA database with histologically confirmed LGG or GBM. Patients were selected based on the availability of high-quality T1-weighted (T1-WI), T1-weighted post-contrast (T1-CE), T2-weighted (T2-WI) and FLAIR images associated with corresponding immune cell markers and clinical information (age, gender and overall survival). Other patients with imaging data are available, however have unclear tumor regions in corresponding MRI sequences and/or no immune cell markers, which are required for the prediction tasks of this study. 

Images for these 151 patients were obtained from The Cancer Imaging Archive (TCIA) \cite{ref26}. Patients have been previously de-identified by TCGA/TCIA, and no institutional review board or Health Insurance Portability and Accountability Act approval were required for our study. Note that the TCGA/TCIA scans were obtained from multiple sites, and thus the scanner model, pixel spacing, slice thickness and contrast varies within the selected cohort. We considered these differences by sampling all volumes to a common voxel resolution of 1 mm$^3$ with a total size of 256$\times$256$\times$slices voxels. Additionally, intensities in each volume were normalised to the $[0,1]$ range. Immune cell markers ($n$=22) are the immune cellular fraction that was estimated using CIBERSORT \cite{ref25}. These proportions were multiplied by leukocyte fraction to yield corresponding estimates in terms of overall fraction in tissue. More details on these markers are reported in \cite{ref7}. Demographic information of the study population can be found in Table \ref{table1}.

\subsection{Segmentation}

Tumors were labeled semi-automatically using the 3D Slicer software 3.6\footnote{\url{http://www.slicer.org/}}. ROIs were determined separately by two expert oncologists, slice by slice from axial images under blind conditions. Those ROIs (contours) were then interpolated to obtain the 3D volumetric tumor mask. The same segmentation procedure was applied to all MRI sequences. 

\subsection{Proposed deep imaging features}

Convolutional neural networks (CNNs) are widely used in medical image analysis and have achieved state-of-art performance for various image classification tasks, in particular when large sets of images are available \cite{ref27,ref28}. Typical CNN architectures are comprised of a repeated stack of convolution and pooling layers, followed by one or more fully-connected layers \cite{ref29}. Convolution layers have a filtering function extracting spatial features from the image. While initial layers capture local image patterns, deep layers extract high-level features representing global structure and texture. To add non linearity, non-saturating activation functions such as the Rectified Linear Unit (ReLU) \cite{ref30} are typically used instead of more traditional functions like the sigmoid. Functions like the ReLU help alleviate the vanishing gradient problem when training deep networks with gradient descent in deep networks. Pooling layers (e.g., maximum or average) are typically added after each convolution layer block to reduce the spatial dimension of activation maps and make the network invariant to small image translations. CNNs for classification also have fully-connected layers followed by an output layer (e.g., Softmax) which converts logits into class probabilities. During training, convolutional filters and fully-connected layer weights are updated using the backpropagation algorithm.

Inspired by studies exploring the flow of information in deep neural networks \cite{ref23,ref24,ref35,ref36}, the entropy of CNN activation maps was proposed as a compact description of texture in medical images \cite{ref22,ref32,ref38}. Although these entropy-based features were shown to be predictive of different diseases, they only offer a limited measure of heterogeneity and do not capture the full range of statistics describing the texture of affected tissues. To overcome this limitation, we derive a richer set of texture features from CNN activations. Toward this goal, deep texture features are extracted from a pre-trained 3D CNN architecture which was previously used in \cite{ref38}. This network was trained on multi-site 3D MRI data for binary classification of Alzheimer's using cross entropy loss, as well as stochastic gradient descent optimization with a momentum of 0.9 and learning rate of 0.0005. We generated texture descriptors from gross total resection (tumor ROI) using the pre-trained 3D CNN and encoded the activation maps ($n$=20) in layer 1 and layer 2, respectively, by applying the following 41 quantifier functions: 
\begin{itemize}
\item\textbf{Histogram}: mean, variance, skewness, kurtosis, energy and entropy;
\item\textbf{Texture}: grey level co-occurrence matrix (GLCM) \cite{ref39}, neighborhood gray-tone difference matrix (NGTDM) \cite{ref40} and gray-level zone size matrix (GLSZM) \cite{ref41}. Image intensities of volumetric tumor/ROIs were uniformly resampled to 32 grey-levels before computation to capture more meaningful texture patterns. 
\end{itemize}
The detailed definition of these descriptor functions/features is provided in the supplementary materials of \cite{ref9}. Descriptors derived from the 20 activation maps were averaged and combined with 4 shape features (porosity, fraction dimension, surface-area and volume) to obtain a set of 45 features. Applying this procedure for each image modality (i.e., T1-WI, T1-CE, T2-WI and FLAIR) yielded a total of 180 DRFs (45 features for each of the four modalities). 

\begin{table}[ht!]
\begin{center}
\caption{Demographic information of the study population.}\label{table1}
\begin{footnotesize}
\setlength{\tabcolsep}{35pt}
\renewcommand{\arraystretch}{1}
\begin{tabular}{ccc}
\toprule    
&  & \textbf{\shortstack{TCGA/TCIA \\ ($n$=151)}} \\
\midrule
{\textbf{Gender}} & Male & 86 \\
& Female & 65 \\
\midrule
{\textbf{Grade}} &  LGG & 83 \\
& GBM & 68 \\
\midrule
{\textbf{Age}} & Median (min-max) & 53 (19-84.8) \\
& Average & 51.34 \\
\midrule
{\textbf{Survival}}  & Median (months) & 14.43 \\
& Dead (censored) & 68 (83) \\
\midrule
{\textbf{\shortstack{Immune cells \\ (presence/absence)}}} & B Cells Memory & 102/49 \\
& B Cells Naive & 75/76 \\
& Dendritic Cells Activated & 74/77 \\
& Dendritic Cells Resting & 31/120 \\
& Eosinophils & 45/106 \\
& Macrophages M0 & 43/108 \\
& Macrophages M1 & 112/39 \\
& Macrophages M2 & 151/0 \\
& Mast Cells Activated & 91/60 \\
& Mast Cells Resting & 63/88 \\
& Monocytes & 139/12 \\
& Neutrophils & 120/31 \\
& NK Cells Activated & 107/44 \\
& NK Cells Resting & 74/77 \\
& Plasma Cells & 59/92 \\
& T Cells CD4 Memory Activated & 12/139 \\
& T Cells CD4 Memory Resting & 135/16 \\
& T Cells CD4 Naive & 47/104 \\
& T Cells CD8 & 111/40 \\
& T Cells Follicular Helper & 132/19 \\
& T Cells gamma delta & 29/122 \\
& T Cells Regulatory Tregs & 34/117 \\
\bottomrule
\end{tabular}
\end{footnotesize}
\end{center}
\end{table}

\subsection{Statistical and survival analysis}

\paragraph{Features analysis} Spearman rank correlation ($\rho$) was used to measure the relationship between pairs of features (e.g., DRF and immune cell markers), and the Wilcoxon test to compare between two groups with continuous variables (e.g., low vs. high immune cell markers) \cite{ref42}. To account for multiple comparisons, all p-values obtained from significance testing were simultaneously corrected according to the Holm-Bonferroni method \cite{ref43}. A threshold of p $<$ 0.05 on corrected p-values was used to identify statistically significant features. 

\paragraph{Survival analysis} The log-rank test \cite{ref44} was used to compare between the lower-than-median and higher-than-median survival groups of patients for each feature. We considered the survival as the number of days to death (i.e., censorship=1) for deceased patients, or days to last visit or follow up (i.e., censorship=0) since initial diagnostic otherwise. Once again, differences were considered statistically significant if p $<$ 0.05 after correction. 

\paragraph{Classification} We used the area under the receiver operator characteristic curve (AUC) to assess whether the DRF signature could classify low from high immune cell markers, and separate patients into short and long survival groups. Specifically, we used all the DRFs as the input of the random forest (RF) classifier to classify between two groups of patients (below vs above median value of immune cell marker; shorter vs longer survival groups using the median survival as cut-off). Note that different classifiers could have been considered for this task, however we chose the RF model since it is recommended when training data is limited and it can be used to inspect predictive features that are most important in classification \cite{ref45}. To train the RF model for classifying between shorter and longer survival groups, we applied an imputation technique to censored patients, for which only a lower bound on survival is known. Specifically, censored patients were assigned the average survival of uncensored subjects with a time-to-death greater or equal to their own time of last visit. 

We considered two strategies for measuring the AUC \cite{ref46}: 1) Leave-one-out cross validation (LOOCV) where training images are divided into $n$ samples and, at each iteration, a single sample is put aside for testing and the remaining $n-1$ samples are used to train the RF classifier. The final reported AUC is the mean over $n$ iterations.  2) Single split, where we divide samples randomly into training ($n$=100) and testing ($n$=51) set, train the RF model using the training samples and test the model using the test samples. The reported AUC value is computed on the set of test samples. All our processing/analysis steps were performed using Matlab's Statistics and Machine Learning Toolbox.

\section{Results}

\subsection{Feature related to immune cell markers} 

\begin{figure}[t!]
    \begin{center}    
    \includegraphics[width=1\textwidth]{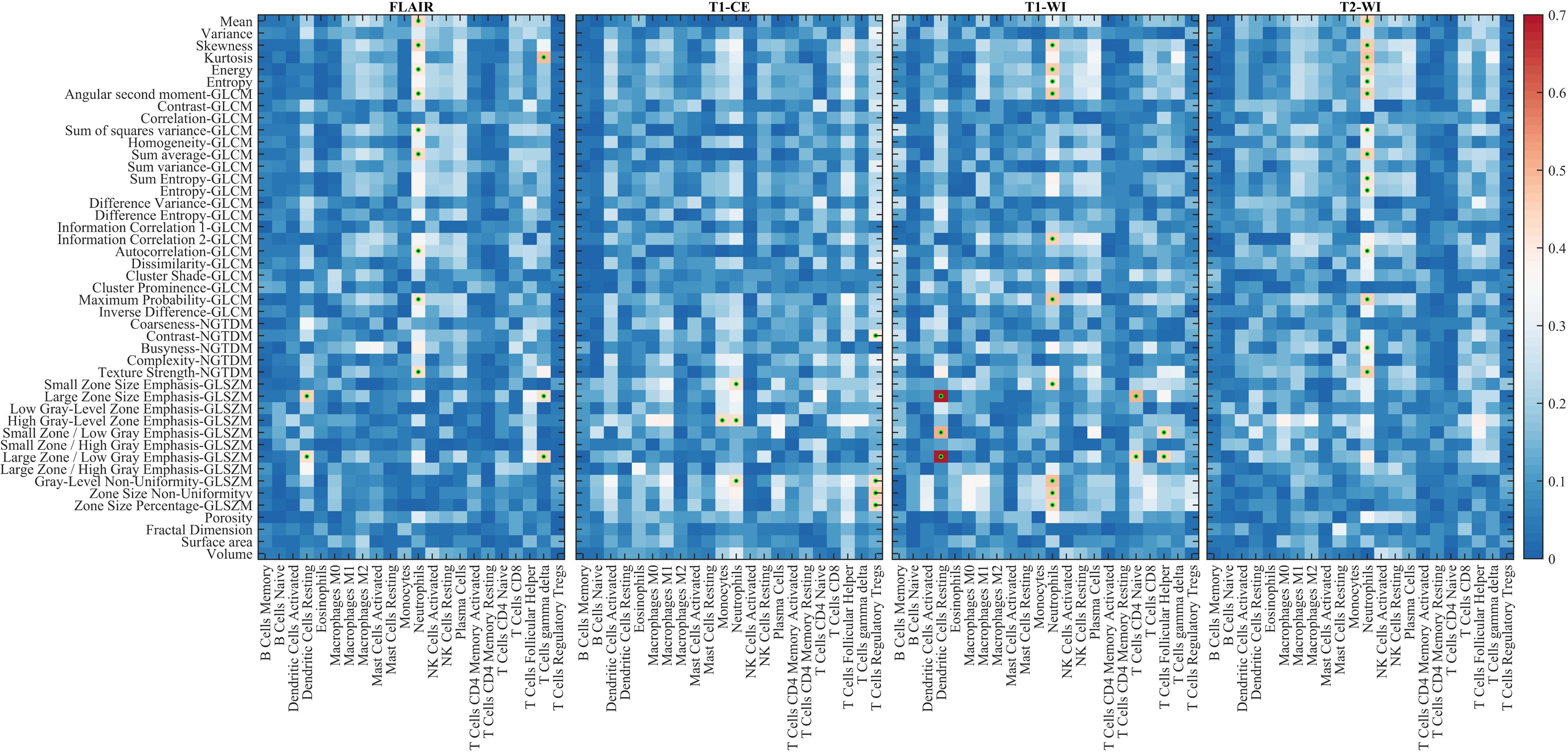}
    \caption{Heatmap of Spearman correlation value between immune cell markers and DRFs. Green circles represent the significant features following Holm-Bonferroni with p $<$ 0.05.}
    \label{fig2}
    \end{center}
\end{figure}

For brain tumor patients in the training set ($n$=100), the analysis on Spearman correlation shows that Dentric-cells-activated (FLAIR and T1-WI), Neutrophils (FLAIR, T1-CE, T1-WI and T2-WI), T-cells-CD8 (FLAIR, and T1-WI), T-Cells-CD4-Naïve (T1-WI) and T-Cells-Regulatory-Tregs (T1-CE) are significantly correlated to DRFs with an absolute value in the range of 0.4-0.74 and corrected p $<$ 0.05. Dentric-cells-activated has the highest absolute correlation of 0.74 with DRFs derived T1-WI (Large Zone Size Emphasis-GLSZM and Large Zone / Low Gray Emphasis-GLSZM) (Figure \ref{fig2}). 

\begin{figure}[ht!]
    \centering
    \includegraphics[width=1\textwidth]{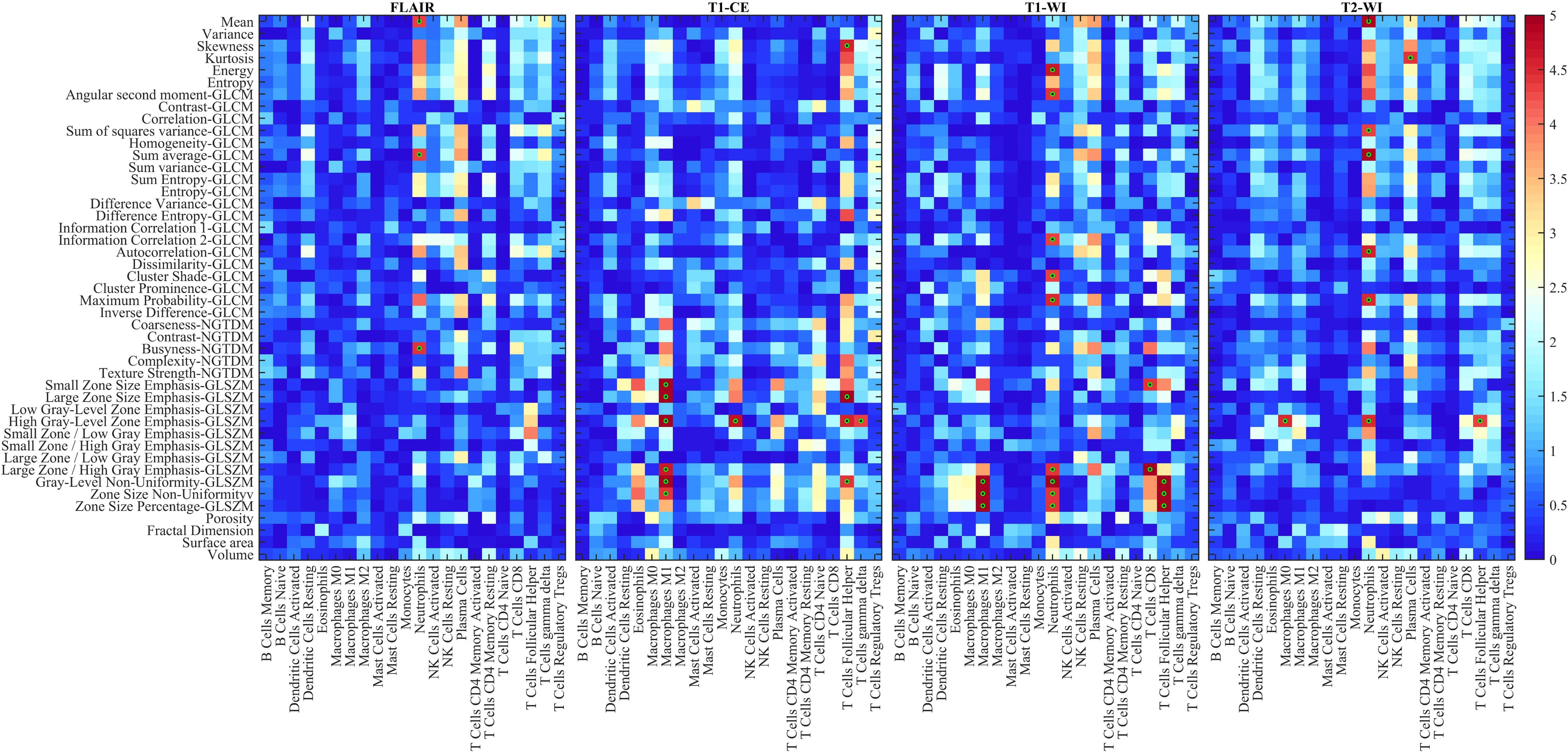}
    \caption{Heatmap of significance value (-$\log_{10}$(p-value)) using the Wilcoxon test to compare DRFs between low and high scores of immune cell markers. Green circles represent the significant features following Holm-Bonferroni with p $<$ 0.05.}
    \label{fig3}
\end{figure}

We then sought to investigate whether individual DRFs could predict immune cell markers. Toward this goal, patients were divided in two groups using the median value of each immune cell markers, i.e. less than median value vs. greater than median value. The ability of each DRF to predict immune cell markers was then evaluated using Wilcoxon significance testing (Figure \ref{fig3}). We see that DRFs are able to discriminate between low and high groups of Neutrophils score (FLAIR, T1-CE, T1-WI and T2-WI), Macrophage-M0 (T2-WI), Macrophage-M1 (T1-CE and T1-WI), T-Cells-Follicular-Helper (T1-CE, T1-WI and T2-WI) and T-Cells-gamma-delta (T1-CE) with corrected p $<$ 0.05. Highest significance values are obtained while using  DRFs extracted from T1-WI (Gray-Level-Non-Uniformity-GLSZM, Zone Size Non-Uniformity and Zone Size Percentage-GLSZM) to compare between low and high scores of Neutrophils, Macrophage-M1 and T-Cells-Follicular-Helper status, respectively. 

Although a direct connection between DRFs and immune cell markers is hard to establish, their association may be explained by cellular characteristics of glioma which affect both the immune system's response and the tumor's appearance in MRI scans. These subtle differences in regional texture can be captured effectively by specifically designed descriptors such as those used in our work.

\subsection{DRFs, clinical and immune cell markers related to survival} 

We next evaluate whether individual features from the set of 45 DRFs, 2 clinical (age and gender) and 22 immune cells features can predict patient survival. For this analysis, we performed a significance test on the same training set of patients with brain tumors ($n$=100). Except for gender, we grouped patients using the median value of each feature (45 DRFs, age, and immune cell markers) to separate patients into groups corresponding to greater vs. less than median feature value. The log-rank significance test was employed to compare between these two groups. We find that age, Macrophages M1, T Cells CD4 Naïve, Neutrophils, T Cells Follicular Helper and 50 DRFs are significantly associated with survival outcome, with the most significant clinical, DRFs and immune cell markers being age, information correlation derived from T1-WI and Macrophage M1, respectively (Table \ref{table2}). Results for all features are reported in Table 2S of supplementary materials.

\begin{figure}[t!]
    \centering
    \includegraphics[width=1\textwidth]{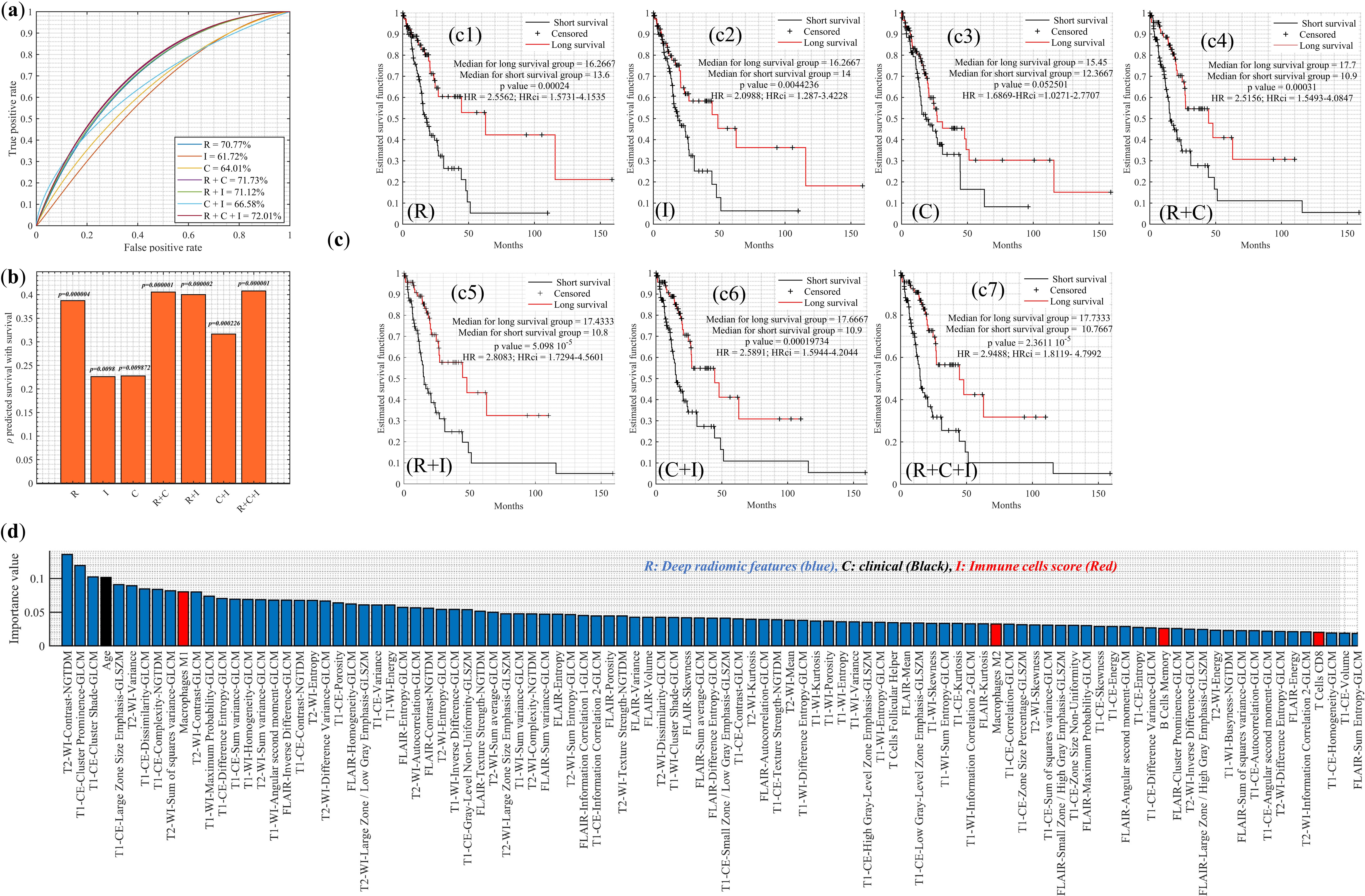}
    \caption{Predicting the survival outcome of 151 patients with brain tumors (LGG+GBM). (\emph{a}) AUC-ROC values using 180 DRFs/radiomic features (R), 2 clinical (C) and 22 immune cell markers (I) and their combinations (R+C, R+I, C+I and R+C+I). (\emph{b}) Bars represent the Spearman correlation between predicted short and long survival score and their corresponding survival. (\emph{c}) Log-rank and Kaplan-Meier estimator (c1-c7) to compare survival between two predicted survival groups based on RF models. (\emph{d}) Importance value of top 100 combined features (R+C+I) from a total 204 features.}
    \label{fig4}
\end{figure}

\subsection{Building radiomic signature related to survival} 

We employed the RF model inside LOOCV to classify the 151 patients in the shorter-term or longer-term survival group using the 180 DRFs (R), 2 clinical features (C), 22 immune cell markers (I) features, or the combination of different feature types (R+I, R+C or R+C+I). Considering individually each type of feature, we see that DRFs give the highest accuracy, with an AUC of 70.77\% compared to 64.01\% for clinical features and 61.72\% for immune cell markers (Figure \ref{fig4}a). Moreover, a significant improvement is achieved when combining both clinical features and immune markers with DRFs, with a highest AUC of 72.01\% when using all features as input to the RF model (R+C+I). Measuring the Spearman correlation ($\rho$) between the score predicted by the RF model and the survival of the 151 patients (Figure \ref{fig4}b), we obtain an absolute correlation of 0.23-0.43 with corrected p $<$ 0.05. Figure \ref{fig4}c shows results of the log-rank significance test and Kaplan-Meier estimator to assess our combined models's ability to predict survival. We observe that differential survival can be predicted when considering the following feature combinations: R (p=0.0002, HR=2.5, CI=1.5-4.1), I (p=0.004, HR=2.09, CI=1.28-3.4), R+C (p=0.0003, HR=2.51, CI=1.54-4.08), R+I (p=5.09$\times$10$^{-5}$, HR=2.8, CI=1.72-4.56), C+I (p=1.9$\times$10$^{-4}$, HR=2.5, CI=1.59-4.2) and R+C+I (p=2.36$\times$10$^{-5}$, HR=2.94, CI=1.81-4.79). We note that combined age and gender features do not lead to statistical significance. Among the 214 features (i.e., 180 DRFs + 22 immune cell markers + 2 clinical), we find 154 features predictive of survival in brain tumor patients (Figure \ref{fig4}d). Specifically, the Contrast-NGTDM from T2-WI, age, and macrophage M1 are the highest predictive features from DRFs, clinical features and immune cell markers, respectively. The Neighbouring Gray Tone Difference Matrix (NGTDM) measures the difference between a gray value and the average gray value of its neighbours within a given distance. A high NGTDM contrast occurs in tumor regions with a large range of gray levels and significant changes in intensity between voxels and their neighbourhood. The importance values of all R+I+C features for predicting survival are reported in Table 4S. 

\begin{figure}[t!]
    \centering
    \includegraphics[width=1\textwidth]{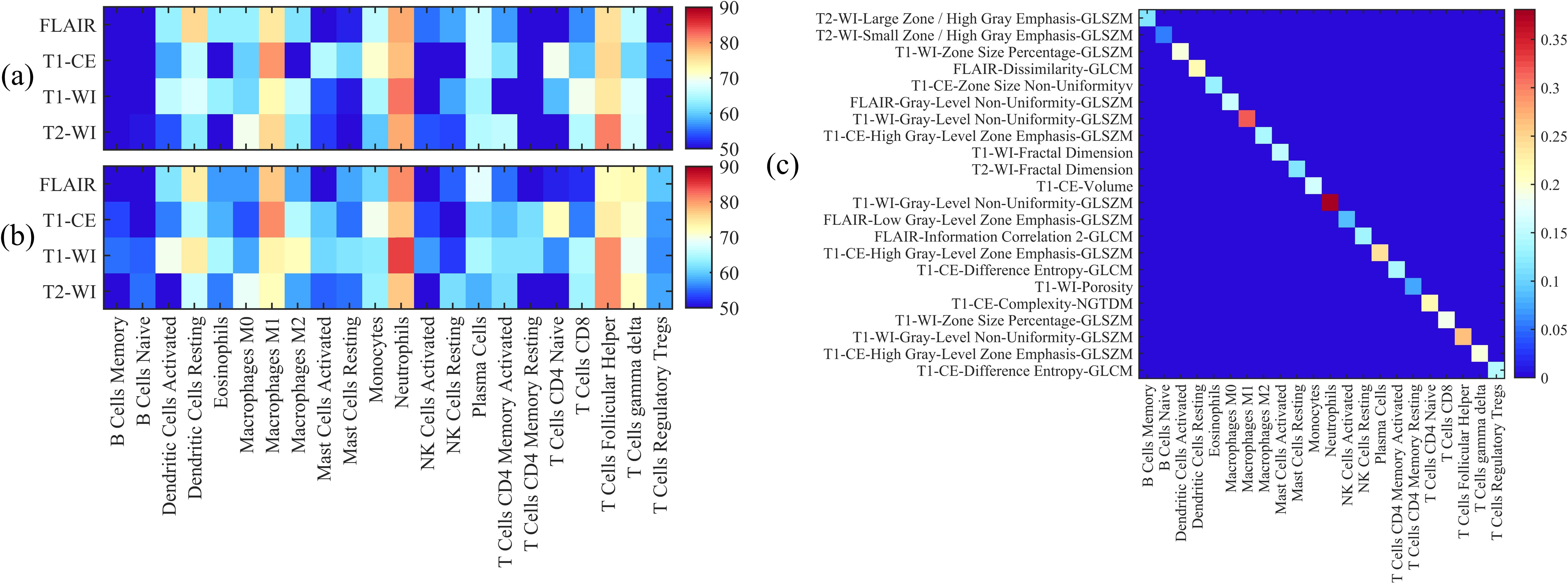}
    \caption{Heatmap of AUC value obtained in predicting low from a high score of 22 immune cell markers using the RF models in (\emph{a}) LOOCV with 100 samples and (\emph{b}) training/testing of 100/51 samples. (\emph{c}) Heatmap of the highest predictive feature for each of the 22 immune cell markers.}
    \label{fig5}
\end{figure}

\subsection{Radiomic signature predicts immune cells status} 

We used the 180 DRFs as input to the RF model from predicting lower and higher values of immune markers, with the LOOCV on the whole set of patients ($n$=151) or with a single training/testing split ($n$=100/51). Results in Figure \ref{fig5} show high AUC values ($>$80\%) for Macrophage M1 (T1-CE), Neutrophils (T1-WI and FLAIR) and T Cells Follicular Helper (T2-WI). These findings are observed for both the LOOCV and training-testing split, with the highest AUC value of 84.81\% obtained while using T1-WI DRFs to predict lower or higher values for the neutrophils marker. Figure \ref{fig5}c shows the most important features for discriminating bettween lower and higher values of each of the 22 immune cell markers. We see that Gray-level-non-uniformity is the most predictive feature for Neutrophils, Macrophage M1 and T cells Follicular Helper markers. Results for all features and cell markers are reported in Table 5S.

\begin{table}[ht!]
\begin{center}
\caption{List of 55 significant features to compare between survival of patient grouped by individual feature values.}\label{table2}
\begin{scriptsize}
\setlength{\tabcolsep}{12pt}
\renewcommand{\arraystretch}{.67}
\begin{tabular}{cccccccc}
\toprule    
\textbf{\shortstack{Features \\ (DRFs, clinical, immune cells)}} & \multicolumn{2}{c}{\textbf{\shortstack{Median \\ survival}}} & \textbf{P-value} &
\textbf{HR} & 
\multicolumn{2}{c}{\textbf{\shortstack{Confidence \\ interval}}} & \textbf{\shortstack{Corrected \\ p-value}} \\
\cmidrule(l{6pt}r{6pt}){2-3}
\cmidrule(l{6pt}r{6pt}){6-7}
 & ${\geq}$ & ${<}$ &  &  & \textbf{CI1} & \textbf{CI2} & \\
\midrule 
Age & 10.6 & 17.0 & 5.45$\times 10^{-8}$ & 4.16 & 2.52 & 6.87 & 0.00001 \\
T1-WI-Information Correlation 2-GLCM & 17.2 & 11.4 & 3.88$\times 10^{-7}$ & 0.27 & 0.16 & 0.44 & 0.00008 \\
T1-WI-Maximum Probability-GLCM & 11.0 & 17.7 & 1.36$\times 10^{-6}$ & 3.51 & 2.14 & 5.76 & 0.00028 \\
Macrophages M1 & 13.3 & 15.5 & 1.40$\times 10^{-6}$ & 3.40 & 2.09 & 5.52 & 0.00028 \\
T2-WI-Sum of squares variance-GLCM & 17.1 & 11.8 & 2.00$\times 10^{-6}$ & 0.30 & 0.18 & 0.48 & 0.00040 \\
T2-WI-Autocorrelation-GLCM & 17.1 & 11.8 & 2.00$\times 10^{-6}$ & 0.30 & 0.18 & 0.48 & 0.00040 \\
T2-WI-Sum variance-GLCM & 17.1 & 11.4 & 2.09$\times 10^{-6}$ & 0.28 & 0.17 & 0.47 & 0.00041 \\
T2-WI-Energy & 11.0 & 18.6 & 3.63$\times 10^{-6}$ & 3.30 & 2.02 & 5.40 & 0.00072 \\
T2-WI-Variance & 16.3 & 11.8 & 4.18$\times 10^{-6}$ & 0.30 & 0.18 & 0.49 & 0.00082 \\
T1-WI-Skewness & 11.9 & 17.4 & 4.42$\times 10^{-6}$ & 3.28 & 2.00 & 5.36 & 0.00086 \\
T1-WI-Energy & 11.0 & 17.7 & 6.69$\times 10^{-6}$ & 3.18 & 1.95 & 5.20 & 0.00130 \\
T2-WI-Entropy-GLCM & 18.4 & 11.4 & 6.80$\times 10^{-6}$ & 0.31 & 0.19 & 0.51 & 0.00131 \\
T1-CE-Large Zone Size Emphasis-GLSZM & 15.4 & 13.1 & 7.49$\times 10^{-6}$ & 0.32 & 0.19 & 0.52 & 0.00144 \\
T2-WI-Entropy & 18.4 & 11.4 & 8.02$\times 10^{-6}$ & 0.32 & 0.19 & 0.52 & 0.00153 \\
T2-WI-Porosity & 11.0 & 18.4 & 8.23$\times 10^{-6}$ & 3.17 & 1.93 & 5.18 & 0.00156 \\
T1-WI-Gray-Level Non-Uniformity-GLSZM & 14.1 & 14.6 & 8.43$\times 10^{-6}$ & 3.07 & 1.90 & 4.97 & 0.00159 \\
T1-WI-Angular second moment-GLCM & 11.0 & 17.7 & 8.64$\times 10^{-6}$ & 3.14 & 1.92 & 5.12 & 0.00162 \\
T2-WI-Mean & 17.1 & 11.8 & 8.81$\times 10^{-6}$ & 0.32 & 0.20 & 0.52 & 0.00165 \\
T2-WI-Sum average-GLCM & 17.1 & 11.8 & 8.81$\times 10^{-6}$ & 0.32 & 0.20 & 0.52 & 0.00165 \\
T2-WI-Small Zone / Low Gray Emphasis-GLSZM & 12.2 & 16.0 & 8.91$\times 10^{-6}$ & 3.20 & 1.94 & 5.28 & 0.00165 \\
T2-WI-Skewness & 11.0 & 17.4 & 9.28$\times 10^{-6}$ & 3.11 & 1.91 & 5.08 & 0.00171 \\
T1-CE-Kurtosis & 15.1 & 14.1 & 9.37$\times 10^{-6}$ & 0.32 & 0.20 & 0.52 & 0.00172 \\
T2-WI-Kurtosis & 11.0 & 17.2 & 9.70$\times 10^{-6}$ & 3.08 & 1.90 & 5.00 & 0.00177 \\
T2-WI-Contrast-NGTDM & 16.3 & 12.0 & 1.02$\times 10^{-5}$ & 0.32 & 0.20 & 0.52 & 0.00184 \\
T2-WI-Sum Entropy-GLCM & 18.4 & 11.4 & 1.03$\times 10^{-5}$ & 0.32 & 0.20 & 0.53 & 0.00186 \\
T1-WI-Entropy & 17.7 & 11.8 & 1.48$\times 10^{-5}$ & 0.33 & 0.20 & 0.54 & 0.00266 \\
T1-CE-Dissimilarity-GLCM & 11.0 & 16.2 & 1.64$\times 10^{-5}$ & 3.08 & 1.87 & 5.07 & 0.00292 \\
T2-WI-Contrast-GLCM & 17.1 & 12.6 & 1.72$\times 10^{-5}$ & 0.33 & 0.20 & 0.54 & 0.00305 \\
T1-CE-Complexity-NGTDM & 11.0 & 15.5 & 1.89$\times 10^{-5}$ & 3.01 & 1.84 & 4.92 & 0.00332 \\
T2-WI-Difference Variance-GLCM & 16.3 & 12.1 & 1.95$\times 10^{-5}$ & 0.33 & 0.20 & 0.54 & 0.00340 \\
T2-WI-Angular second moment-GLCM & 11.9 & 17.7 & 2.23$\times 10^{-5}$ & 2.96 & 1.82 & 4.81 & 0.00388 \\
T1-WI-Sum Entropy-GLCM & 17.7 & 11.8 & 2.31$\times 10^{-5}$ & 0.34 & 0.21 & 0.55 & 0.00400 \\
T1-CE-Skewness & 15.1 & 14.0 & 3.30$\times 10^{-5}$ & 0.34 & 0.21 & 0.56 & 0.00567 \\
T2-WI-Maximum Probability-GLCM & 11.9 & 17.4 & 3.38$\times 10^{-5}$ & 2.88 & 1.77 & 4.68 & 0.00579 \\
T2-WI-Volume & 17.8 & 11.4 & 3.42$\times 10^{-5}$ & 0.35 & 0.21 & 0.56 & 0.00581 \\
T2-WI-Large Zone / Low Gray Emphasis-GLSZM & 11.7 & 17.7 & 5.80$\times 10^{-5}$ & 2.79 & 1.72 & 4.53 & 0.00980 \\
T1-WI-Porosity & 12.0 & 17.7 & 5.82$\times 10^{-5}$ & 2.83 & 1.73 & 4.64 & 0.00980 \\
T1-CE-Gray-Level Non-Uniformity-GLSZM & 11.7 & 15.4 & 6.84$\times 10^{-5}$ & 2.79 & 1.71 & 4.56 & 0.01142 \\
T1-CE-Contrast-NGTDM & 12.2 & 15.6 & 7.42$\times 10^{-5}$ & 2.80 & 1.71 & 4.58 & 0.01231 \\
T1-WI-Small Zone Size Emphasis-GLSZM & 14.1 & 14.6 & 8.14$\times 10^{-5}$ & 2.70 & 1.67 & 4.36 & 0.01343 \\
FLAIR-Kurtosis & 12.2 & 17.4 & 8.33$\times 10^{-5}$ & 2.71 & 1.67 & 4.39 & 0.01367 \\
FLAIR-Difference Entropy-GLCM & 16.8 & 11.4 & 1.10$\times 10^{-4}$ & 0.37 & 0.23 & 0.61 & 0.01791 \\
T Cells CD4 Naive & 15.3 & 13.6 & 1.21$\times 10^{-4}$ & 0.36 &  0.22 & 0.59 & 0.01963 \\
T2-WI-Large Zone Size Emphasis-GLSZM & 11.9 & 16.5 & 1.31$\times 10^{-4}$ & 2.67 & 1.64 & 4.34 & 0.02113 \\
Neutrophils & 14.7 & 13.1 & 1.46$\times 10^{-4}$ & 2.60 & 1.61 & 4.20 & 0.02338 \\
FLAIR-Energy & 12.2 & 16.9 & 1.49$\times 10^{-4}$ & 2.61 & 1.61 & 4.22 & 0.02376 \\
FLAIR-Angular second moment-GLCM & 12.2 & 16.9 & 1.59$\times 10^{-4}$ & 2.60 & 1.61 & 4.20 & 0.02511 \\
FLAIR-Mean & 17.1 & 12.1 & 1.61$\times 10^{-4}$ & 0.38 & 0.24 & 0.62 & 0.02532 \\
FLAIR-Sum average-GLCM & 17.1 & 12.1 & 1.61$\times 10^{-4}$ & 0.38 & 0.24 & 0.62 & 0.02532 \\
T1-CE-Large Zone / High Gray Emphasis-GLSZM & 15.7 & 12.0 & 1.75$\times 10^{-4}$ & 0.38 & 0.23 & 0.62 & 0.02719 \\
FLAIR-Maximum Probability-GLCM & 12.0 & 17.4 & 1.83$\times 10^{-4}$ & 2.59 & 1.60 & 4.19 & 0.02813 \\
FLAIR-Texture Strength-NGTDM & 11.7 & 16.5 & 2.06$\times 10^{-4}$ & 2.56 & 1.58 & 4.13 & 0.03156 \\
T Cells Follicular Helper & 12.2 & 15.2 & 2.61$\times 10^{-4}$ & 2.53 & 1.56 & 4.11 & 0.03969 \\
FLAIR-Skewness & 12.2 & 16.5 & 2.82$\times 10^{-4}$ & 2.50 & 1.55 & 4.04 & 0.04252 \\
FLAIR-Sum Entropy-GLCM & 17.7 & 11.4 & 3.23$\times 10^{-4}$ & 0.40 & 0.25 & 0.65 & 0.04844 \\
\bottomrule
\end{tabular}
\end{scriptsize}
\end{center}
\end{table}

\section{Discussion}

The prediction of immune status may help identify cancer patients that will respond to treatment \cite{ref47}. Radiomics is a non-invasive technique for the automated prognosis of various types of tumors which uses a wide range of imaging features extracted from a region of interest (ROI) \cite{ref48,ref49}. While standard radiomic features have shown promising results \cite{ref9,ref13,ref16,ref18,ref50}, the use of multiscale features from different 3D CNN layers as learnable radiomics descriptors remains limited \cite{ref38}. Using these deep features in combination with clinical variables and immune cell markers could improve the prediction of treatment response, thereby enabling the selection of optimal treatment for individual patients. Toward this goal, we proposed a novel radiomic analysis pipeline that considers 41 descriptors extracted from 3D CNN activation maps to predict survival and the value of immune cell markers. 

Our study showed that DRFs derived from a pre-trained 3D CNN can accurately predict the survival group of patients as well as the low or high value of immune cell markers. In our experiments, the combination of DRFs, clinical features and immune cell markers achieved the highest AUC of 72.01\% (corrected p $<$ 0.05) for classifying patients in groups corresponding to shorter (i.e., below median) or longer (i.e., above median) survival. The most predictive features in the RF model were found to be DRFs, age and Macrophage M1 (Figure \ref{fig4}). This result is consistent with previous work in the literature which showed age \cite{ref51,ref52}, Macrophage \cite{ref7,ref53} and radiomic features to be associated with survival outcome \cite{ref50,ref54,ref55,ref56}.
The proposed analysis also shown that combined DRFs give the highest accuracy to predict the low or high value of Macrophage-M1, Neutrophils and T-cells-follicular-helper immune cell markers with an AUC $>$80\%. Moreover, the gray-level non uniformity descriptor derived from T1-WI images was the most predictive feature for these three immune cell markers (Figure \ref{fig5}). This confirms the potential of radiomic features for predicting immune markers \cite{ref50,ref57,ref58}.

By incorporating immune cell markers in the proposed method, this study aims to enhance the impact of immunotherapy in clinical practice. This is similar to the recent work in \cite{ref16} which presented a radiomics approach to predict the response to cancer immunotherapy. Our study is also related to radiogenomics profiling methods to identify MRI-associated immune cell markers in GBM which are correlated with prognosis. For instance, CD49d expression level was found to be correlated with apparent diffusion coefficient (ADC) and shown to be a useful biomarker to predict progression of GBM patients \cite{ref59}. Likewise, the reduction in the tumor-promoting effects of monocytes/macrophages in GBM was considered as an adjuvant treatment for glioma \cite{ref60}. It was demonstrated that determining the various roles of immune cell markers has an impact on the diagnosis and prediction of cancer progression \cite{ref61} and recurrent GBM \cite{ref62,ref63}.

Our findings suggest the proposed imaging features to be related to neutrophils, macrophage and follicular helper T-cells markers. Pre-treatment neutrophils can potentially serve as a prognostic marker in predicting the chemotherapeutic response and survival outcomes in glioma \cite{ref64}. Macrophage and follicular helper T-cells were also demonstrated as immune therapeutic markers for glioma patients and were correlated with an unfavorable prognostic \cite{ref65, ref66, ref67}. Moreover, radiomic signatures (i.e., imaging features) were shown to be related to tumor variations and changes in vascularization and inflammatory status \cite{ref68, ref69, ref70}. However, a systematic biological approach such as an animal model is necessary to clarify and validate the relationship between immune cell markers and proposed imaging features.

Our study has some limitations worth of mention, such as the use of standard MR imaging sequence (T1-WI, T1-CE, FLAIR and T2-WI). Thus, considering ADC and dynamic contrast-enhanced (DCE) sequences with additional immune cell markers could improve our method's accuracy. Furthermore, validating our method on other datasets could further help demonstrate its usefulness to clinical practice. 

\section{Conclusion}

This study investigated deep radiomic descriptors derived from 3D CNN features maps for predicting the immune cell makers and survival outcome in patients with brain tumors. Our findings suggest associations between deep features, immune cell markers and survival of patients with glioma. Our work provides a model that could potentially be used for accurately predicting three immune cell markers and the survival outcome of patients. Motivated by these results, we aim to expand the proposed deep radiomic pipeline across various cancer types.



\appendix

\end{document}